\documentclass[]{bytedance_seed}

\usepackage{minitoc}
\usepackage{url}
\usepackage{amssymb}
\usepackage[utf8]{inputenc}
\usepackage{microtype}
\usepackage{booktabs}
\usepackage{pifont} 
\usepackage{multirow}
\usepackage{makecell}
\usepackage{paralist}
\usepackage{xspace}
\usepackage{color}
\usepackage{xcolor}
\usepackage{colortbl}
\usepackage{adjustbox}
\usepackage{hyperref} 
\usepackage[edges]{forest}
\usepackage{tikz} 
\usepackage{caption}
\usepackage{amsfonts}
\usepackage[toc,page,header]{appendix}
\usepackage[utf8]{inputenc}
\usepackage{bbm}
\usepackage{enumitem}
\usepackage{mathtools}

\definecolor{seedc}{RGB}{7, 92, 173}

\newcommand{\name}[1]{GR-RL}

\newcommand{\hardware}[1]{ByteMini}

\newcommand{\ba}{\mathbf{a}}

\renewcommand{\paragraph}[1]{\vspace{0.1em}\noindent\textbf{#1}}

\title{GR-RL: Going Dexterous and Precise for Long-Horizon Robotic Manipulation}

\author[]{ByteDance Seed}
\correspondence{\email{xiao.ma@bytedance.com}}

\contribution[]{Full Author List in \hyperref[sec:contributions]{Contributions and Acknowledgments}}

\abstract{

We present \name{}, a robotic learning framework that turns a generalist vision-language-action (VLA) policy into a highly capable specialist for long-horizon dexterous manipulation.
Assuming the optimality of human demonstrations is core to existing VLA policies. However, we claim that in highly dexterous and precise manipulation tasks, human demonstrations are noisy and suboptimal.
\name{} proposes a multi-stage training pipeline that filters, augments, and reinforces the demonstrations by reinforcement learning.
First, \name{} learns a vision-language-conditioned \textit{task progress}, filters the demonstration trajectories, and only keeps the transitions that contribute positively to the progress. Specifically, we show that by directly applying offline RL with sparse reward, the resulting $Q$-values can be treated as a robust progress function.
Next, we introduce \textit{morphological symmetry augmentation} that greatly improves the generalization and performance of \name{}.
Lastly, to better align the VLA policy with its deployment behaviors for high-precision control, we perform online RL by learning a latent space noise predictor. 
With this pipeline, GR-RL is—to our knowledge—the first learning-based policy that can autonomously lace up a shoe by threading shoelaces through multiple eyelets with an 83.3\% success rate, a task requiring long-horizon reasoning, millimeter-level precision, and compliant soft-body interaction.
We hope GR-RL provides a step toward enabling generalist robot foundation models to specialize into reliable real-world experts.

}

\checkdata[Project Page]{\url{https://seed.bytedance.com/gr_rl}}

\begin{document}
\maketitle

\section{Introduction}
\label{sect:intro}

The emergence of large-scale vision-language-action (VLA) models has rapidly advanced the ambition of building generalist robotic agents capable of performing a broad range of tasks given visual observations and free-form natural language instructions.
Recent systems~\cite{brohan2022rt, brohan2023rt, black2410pi0, team2025gemini, bjorck2025gr00t, cheang2024gr, barreiros2025careful} have demonstrated impressive generalization across objects, environments, and semantic concepts, suggesting that robotics can benefit from the scaling laws that propelled progress in vision-language models. 

However, being general is not equivalent to being reliable, and current VLA policies still fall short in two fundamental aspects for real-world deployment: (1) \textit{Dexterity with precision} – millimeter-level control over deformable objects remains unsolved.
(2) \textit{Long-horizon robustness} – errors accumulate over steps, and it gets worse when coupled with high-precision dexterous manipulation.
Consider the task of threading the shoelaces: (1) the robot should be dexterous enough to handle the deformable objects, including both the shoelace and the shoe; (2) the robot needs millimeter-level control precision to thread the shoelace into the eyelets; (3) the robot also needs long-horizon manipulation capabilities to handle diverse and unexpected scenarios. Classic methods tackle shoelacing by motion planning with predefined action primitives and designed patterns~\cite{luo2023shoelacing,luo2024benchmarking,luo2025tsl}, and consequently, generalization to unseen configurations, recovering from failures, and other dexterous skills remain an open question. Naive extensions to behavior cloning will result in sub-optimal and limited skills in shoelacing~\cite{luo2025interface}.


Our starting point is GR-3~\cite{cheang2025gr}, a large-scale VLA policy trained from internet data, robot trajectories, and human demonstrations. Despite its strong generalization capabilities, GR-3 would fail when precision, dexterity, and long-horizon robustness matter. We observe that there are two key bottlenecks: (1) \textit{suboptimal human demonstrations} and (2) the \textit{demonstration and inference mismatch}.
Under extreme precise and dexterous manipulation scenarios, human demonstrators would slow down, hesitate, and introduce noisy suboptimal demonstrations to the policy. In addition, during standard offline training, a VLA policy learns to mimic human demonstrators by predicting action chunks of \textit{fixed lengths} derived from a sliding window given human demonstrations~\cite{zhao2023learning,chi2024diffusionpolicy}. However, to achieve smooth inference and control, post-smoothing to predicted trajectories (e.g., temporal ensembling~\cite{zhao2023learning}), asynchronous receding horizon control~\cite{chi2024diffusionpolicy,black2410pi0,intelligence2025pi_}, and other control-level optimizations are often applied. These system-level optimization methods are necessary for the smooth execution of a learning-based policy, while inevitably causing the mismatch between model training and inference.

\begin{figure}
    \centering
    \includegraphics[width=\linewidth]{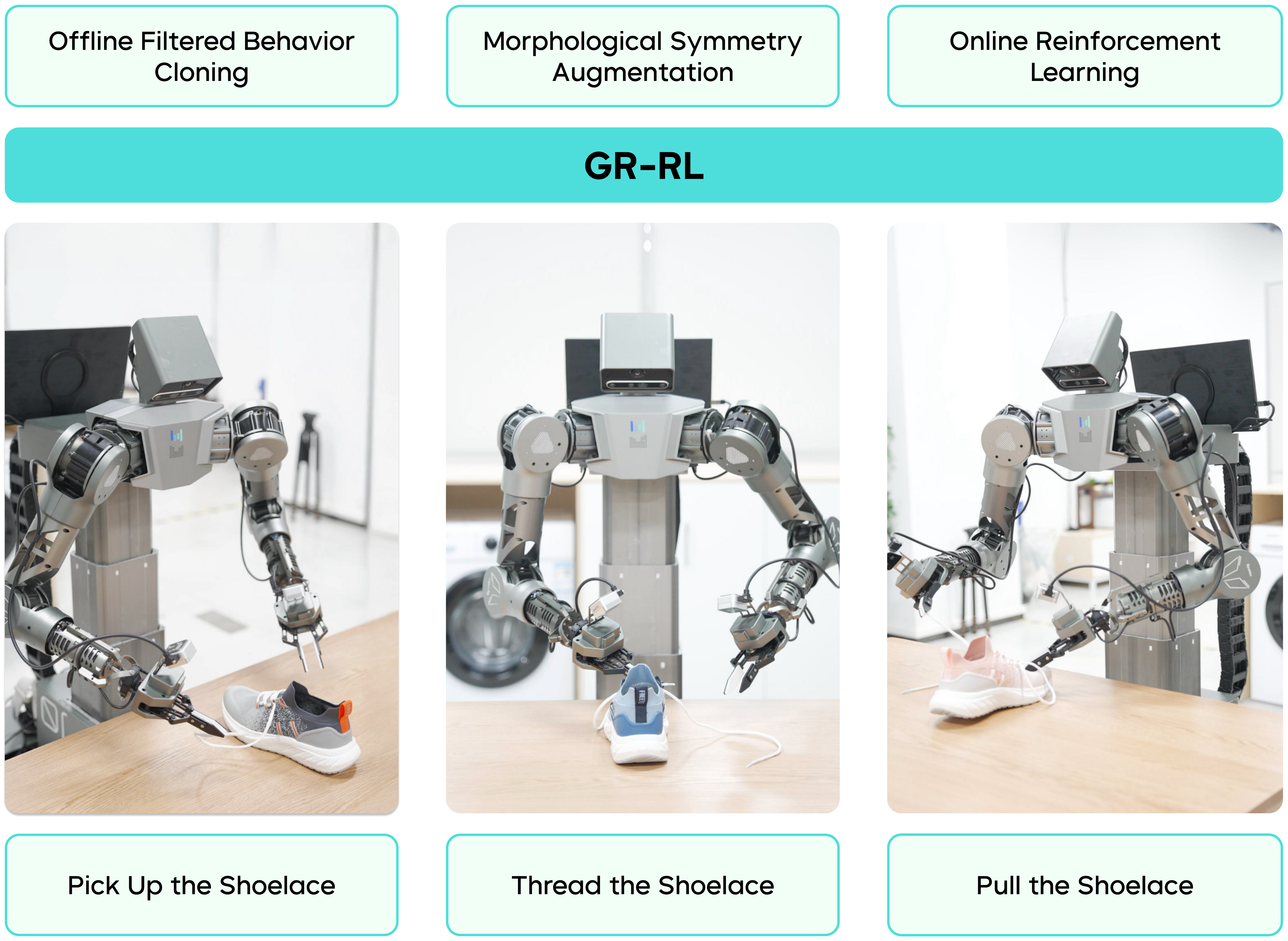}
    \caption{\name{} performs long-horizon, dexterous, and high-precision manipulation, in the task of shoe lacing, by adopting a multi-stage training pipeline, consisting of 1) offline filtered behavior cloning with learned task progress, 2) simple yet effective action augmentation, 3) online reinforcement learning.}
    \label{fig:main}
\end{figure}

We present \name{} for long-horizon dexterous and precise manipulation.
\name{} adopts a multi-stage reinforcement-augmented training pipeline that filters, augments, and reinforces the suboptimal and mismatched human demonstrations.
First, instead of directly running behavior cloning on the entire human demonstration dataset, we initialize the base \name{} VLA policy by cloning the \textit{filtered} trajectories. Specifically, we train a critic model on both successful and failed trajectories with offline reinforcement learning (RL)~\cite{fujimoto2021minimalist}.
Given a sparse reward at the end of the episode, the predicted value naturally reflects the progress of the task, which we further use to filter only transitions that contribute positively to the progress and discard the rest. 
We adopt distributional critics and observe that they give much more robust performance under offline sparse reward scenarios.
Next, initialized from the offline pretrained checkpoints, we perform online reinforcement learning to further explore and fix the failure modes of the base policy. In particular, we achieve this by learning to steer the denoising process towards high-return regions~\cite{wagenmaker2025steering}. 
Lastly, we devise a simple yet effective method to augment the robot actions by mirroring the robot actions and observations, with a flipped text description. Such a scheme drastically improves the overall success rate and generalization capabilities of our policy.

To the best of our knowledge, \name{} is the first learning-based model that can thread the shoelace through multiple eyelets, achieving an overall 83.3\% success rate. We hope \name{} provides a step towards enabling generalist robot foundation models to specialize and be applicable in challenging real-world scenarios.

\begin{figure}[!t]
    \centering
    \includegraphics[width=\linewidth]{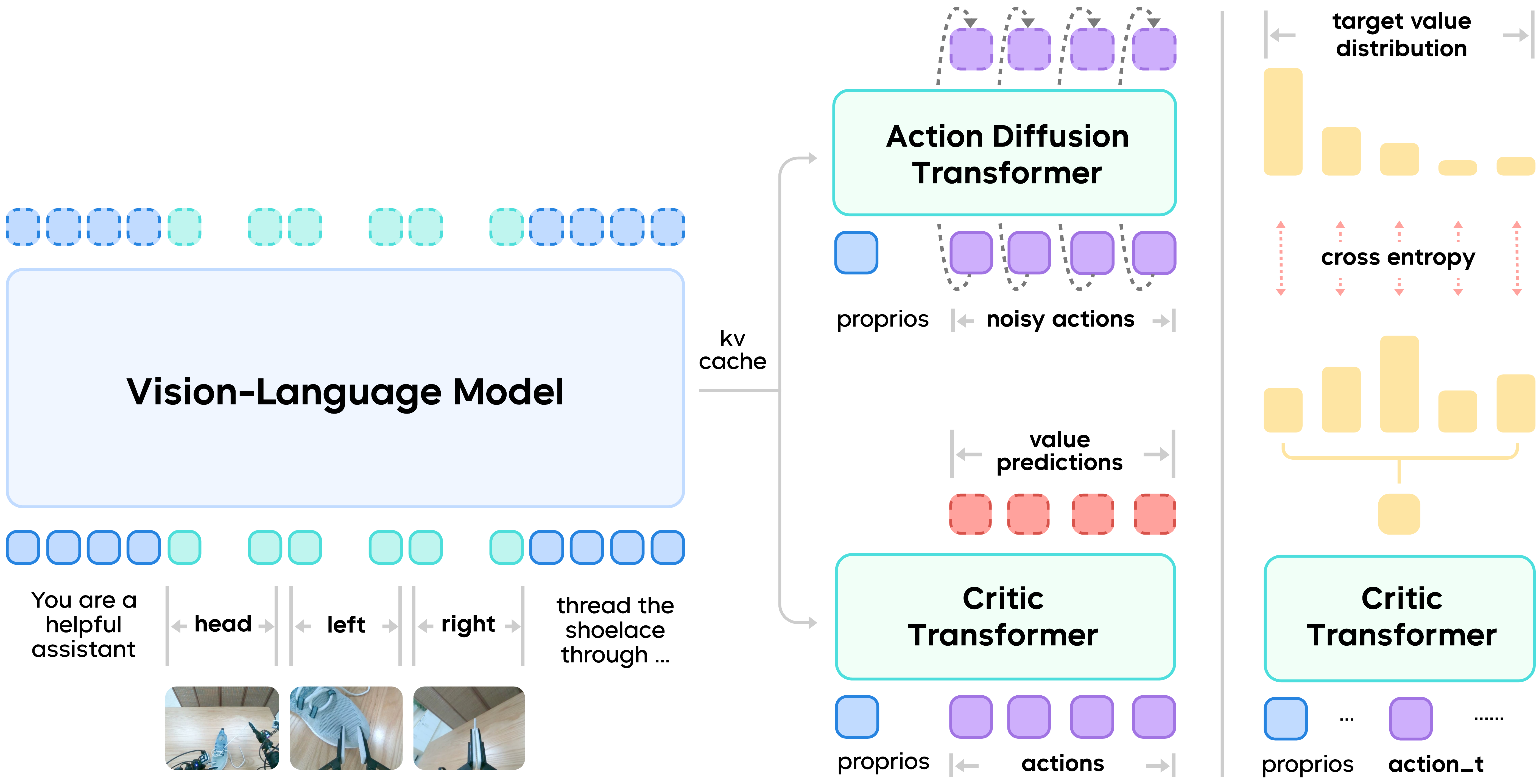}
    \caption{\textbf{The \name{} Model.} \name{} adopts a Mixture-of-Transformer (MoT) architecture. It is co-trained on robot vision-language-action trajectories via a flow-matching objective, and Temporal-Difference (TD) errors via distributional reinforcement learning.
    }
    \label{fig:arch}
\end{figure}

\section{The \name{} Model}
\label{sec:gr_model}
\name{} adopts a Mixture-of-Transformer architecture, consisting of a vision-language-action (VLA) model $\pi_{\theta}$ and a multi-task critic $Q_\phi$ with a total of 5B parameters.

\paragraph{Policy:}
$\pi_\theta$ controls a bi-manual robot with a mobile base by generating a $k$-length action chunk $\ba_t  = a_{t:t+k}$ conditioned on the input language instruction $l$, observation $\mathbf{o}_{t}$, and robot state $\mathbf{s}_{t}$, \textit{i.e.}, $\mathbf{a}_{t} = \pi_{\theta}(l, \mathbf{o}_{t}, \mathbf{s}_{t})$.
Following the architecture design of GR-3~\cite{cheang2025gr}, \name{} uses Qwen2.5-VL-3B-Instruct~\cite{bai2025qwen2} as the Vision-Language-Model (VLM) backbone, and predicts the action chunks $\mathbf{a}_t$ with an action diffusion transformer (DiT) trained by flow matching objectives~\cite{black2410pi0, liu2022rectified, lipman2022flow}. Specifically, we follow GR-3 and use only the KV cache from the latter half of the VLM layers for fast inference.

\paragraph{Critic:}
Similar to the policy $\pi_\theta$, $Q_\phi(\mathbf{o}_t, l, \mathbf{s}_t, \mathbf{a}_t)$ is a causal transformer that evaluates each action. Specifically, we follow $Q$-chunking~\cite{li2025reinforcement,seo2024coarse,li2025toperl} and predict a chunk of $Q$-values for each action chunk $\ba_t$ and adopt distributional reinforcement learning~\cite{bellemare2017distributional,hessel2018rainbow,seo2024coarse,farebrother2024stop}. 
Different from unbounded regression-based policy evaluation, distributional critics treat values as a discrete distribution with upper and lower bounds. 
This naturally captures uncertainty in real-world trajectories.
Under sparse reward settings, distributional critics give much stronger robustness than the non-distributional ones. By setting the upper bound to 1 and the lower bound to 0, our learned critic naturally reflects the \textit{progress} of the task, as shown in Fig.~\ref{fig:method:progress}.

\section{Training Recipe}
\label{sec:data_training}
Human demonstrations are suboptimal and noisy. In the context of long-horizon dexterous high-precision manipulation, human demonstrators tend to hesitate, make mistakes, and try to finish the task with inconsistent behaviors. In addition, inference-time optimizations, such as whole-body receding horizon control and temporal ensembling~\cite{zhao2023learning,chi2024diffusionpolicy}, introduce further mismatch between training and deployment, which exacerbate the negative effect of suboptimal demonstrations.

We introduce a reinforcement-augmented training pipeline to achieve dexterous and precise robotic manipulation from human demonstrations. To prevent the policy from memorizing suboptimal behaviors during supervised learning, we learn a task progress model using offline RL and use it to filter out detrimental data (Sec.~\ref{sec:training:filter}). We then augment the demonstrations to improve the robustness of the offline policy based on the symmetry in bimanual manipulation (Sec.~\ref{sec:training:mirror}). Finally, we perform online reinforcement learning, allowing the model to learn from trial-and-error in a closed loop, which mitigates the mismatch between training and deployment and boosts its overall performance (Sec.~\ref{sec:training:online}).  

\subsection{Data Filtering with a Learned Task Progress Evaluator}\label{sec:training:filter}
For high-precision manipulation with deformable objects, such as shoe lacing, collecting perfect demonstrations is extremely difficult. Even the trajectories collected by experienced teleoperators contain suboptimal fragments: erroneous attempts, hesitations, etc. Directly imitating all the data would unnecessarily introduce noisy multi-modal actions to the training and lead to a policy with suboptimal performance. However, labeling the suboptimal fragments is non-trivial and might introduce even more subjective and noisy human priors.
To identify and filter out suboptimal actions, we propose to learn a task progress model using offline RL. Specifically, we train the critic using TD3+BC~\cite{fujimoto2021minimalist}.
We adopt a sparse reward defined as
\begin{equation}
    r(\mathbf{o}_t, l, \mathbf{s}_t, \mathbf{a}_{t}) = 
    \begin{cases}
        \gamma^{T-t}\mathbb{I}(\tau), & t > T-k, \\
        0, & t \leq T-k,
    \end{cases}
\end{equation}
where $\mathbb{I}(\cdot)$ is an indicator function to evaluate whether a trajectory $\tau$ is successful or not, $T$ is the length of the trajectory, and $\gamma$ means the discount factor. 
Since most of the collected trajectories end in success, we annotate the retry keyframes in each demonstration and create more failed trajectories in hindsight~\cite{NIPS2017_453fadbd}.
Suppose frames $m_i, 0\leq i< M$ are marked as retry keyframes in a successful trajectory $\tau_{0:T}$, we can augment $M$ failed trajectories $\tau_{0:m_i}, 0\leq i<M$ in addition to the original successful one. With temporal difference learning over both successful and failed data, the critic $Q_{\phi}$ could function as a robust task progress evaluator.
\begin{figure}[p]
    \centering
    \includegraphics[width=0.98\linewidth]{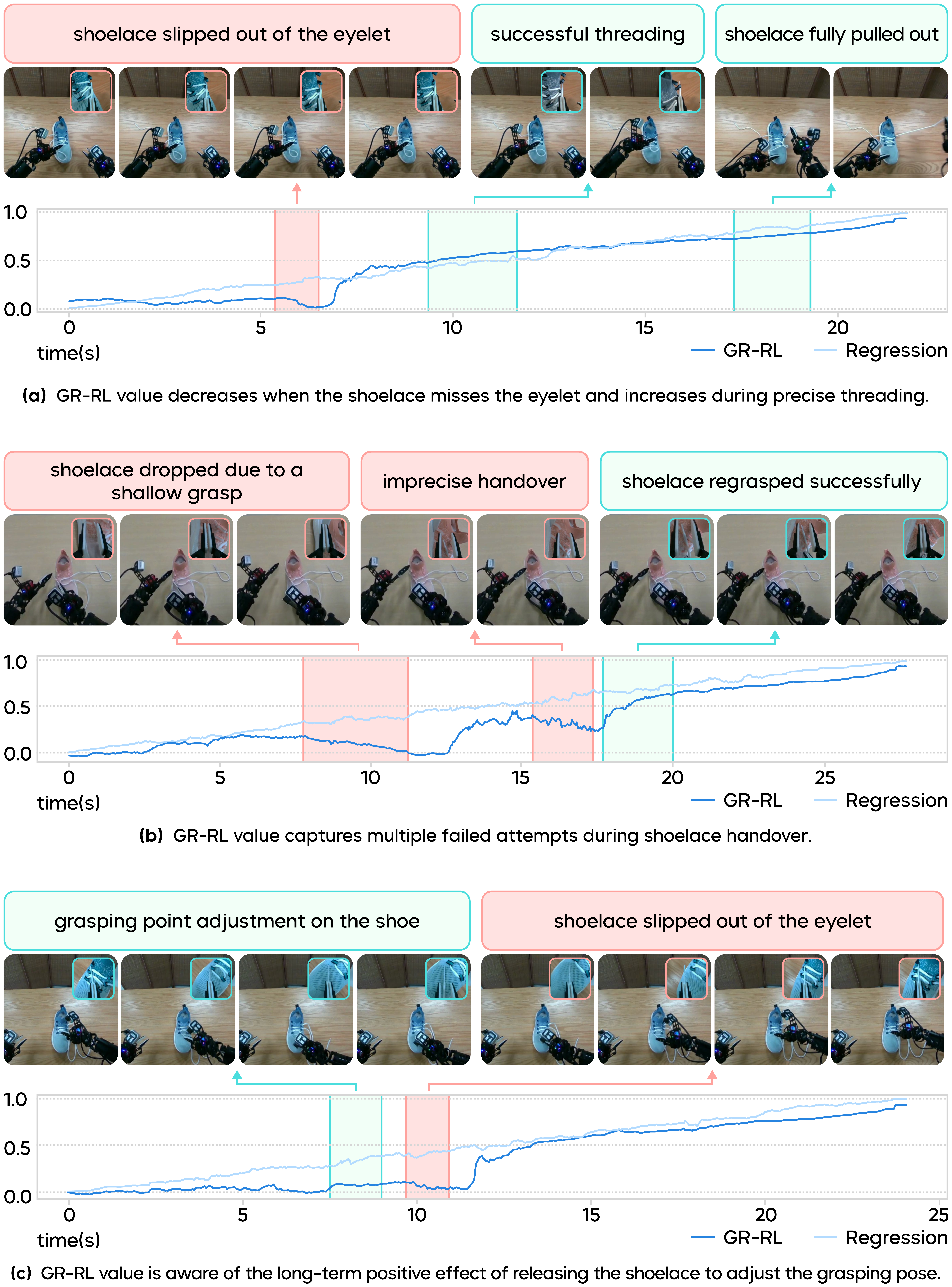}
    \caption{Examples of learned task progress.}
    \label{fig:method:progress}
\end{figure}
After obtaining a task progress model, we evaluate $Q_{\phi}$ and compute its mean value of the categorical distribution as progress $\rho$ for all the transitions in the dataset, 
\begin{equation}
    \rho_t \coloneq \mathtt{mean}(Q_{\phi}(\mathbf{o}_t, l, \mathbf{s}_t, \mathbf{a}_t)).
\end{equation}
An example of the predicted progress is shown in Fig.~\ref{fig:method:progress}. We can observe a sudden drop in the progress when the teleoperator makes a mistake.  
We define a sample $(\mathbf{o}_t, l, \mathbf{s}_t, \mathbf{a}_t)$ at timestep $t$ as suboptimal if there is a value drop greater than a certain threshold $\delta$ in the sequence $\rho_{t:t+k}$, and exclude all the suboptimal ones from the dataset for policy learning. We can then train  $\pi_{\theta}$ simply with behavior cloning using the filtered dataset of higher quality. 

\subsection{Imitation Learning with Data Augmentation}\label{sec:training:mirror}

During the offline training stage, we apply a simple yet effective \textit{morphological symmetry augmentation} paradigm, which further boosts the policy performance.
The augmentation paradigm leverages the morphological symmetry in our bimanual task settings.
For image observations $\mathbf{o}_t$, we flip all the images horizontally, then swap the images from the left wrist with those from the right wrist. All transformations in proprioception states $\mathbf{s}_t$ and actions $\mathbf{a}_t$ are converted via mirror symmetry in the world frame, and then transformed back to local wrist frames. We also flip the spatial description in the language instructions accordingly, e.g., changing ``the hole on the left'' to ``the hole on the right''. Empirically, the symmetry data augmentation can effectively enhance the performance of the policy.

\subsection{Online Steering for Policy Deployment Alignment}\label{sec:training:online}
System-level postprocessing is commonly applied to ensure smooth robot motions when deploying chunking policies, such as temporal ensembling and receding horizon control~\cite{chi2024diffusionpolicy,zhao2023learning}. 
However, these optimization tricks cause a mismatch between training and deployment: what the policy has seen during training (raw actions) is different from the ones actually being executed during deployment (optimized actions).
In the context of dexterous and precise manipulation, the mismatch becomes non-negligible.
To adapt to the discrepancy, we find it crucial for the model to explore and improve itself via closed-loop online interactions with \textit{aligned} actions. 

Performing online RL in long-horizon, precise manipulation tasks remains non-trivial, especially in exploration. Since the task requires millimeter precision to complete, adding noise to wrist poses or joint positions hardly leads to success. 
We instead perform structured exploration in a latent space and steer the trained flow policy~\cite{wagenmaker2025steering}. Specifically, we add a \textit{noise predictor} $\pi_{\theta^\prime}$ after the shared VLM backbone to predict the initial noise $\mathbf{\epsilon}_t$ for the action DiT. 
The number of trainable parameters in $\pi_{\theta^{\prime}}$ is 51.5M. 
To avoid generating arbitrary actions from noise out of the offline training distribution, we penalize the noise predictor when its output diverges from the original normal distribution beyond a certain threshold $\beta$. Following~\cite{wagenmaker2025steering}, we also distill a $Q$ function over noise space $Q_{\phi^{\prime}}(\mathbf{o}_t, l, \mathbf{s}_t, \mathbf{\epsilon}_t)$ to avoid back-propagating through the flow model during policy optimization. The critic in the original action space $Q_{\phi}(\mathbf{o}_t, l, \mathbf{s}_t, \mathbf{a}_t)$ is trained via standard TD3. The online training objectives for the noise transformer and the critic in noise space are as follows:
\begin{equation}
    \mathcal{L}(\pi_{\theta^{\prime}}) = \mathbb{E}_{(\mathbf{o}_t, l, \mathbf{s}_t)\sim \mathcal{D}} \left[-Q_{\phi^{\prime}}(\mathbf{o}_t, l, \mathbf{s}_t, \mathbf{\epsilon}_t) + c \max( \frac{1}{2}\Vert \mathbf{\epsilon}_t\Vert^2 - \beta, 0)\right], \mathbf{\epsilon}_t\sim \pi_{\theta^{\prime}}(\mathbf{o}_t, l, \mathbf{s}_t),
\end{equation}
\begin{equation}    
    \mathcal{L}(Q_{\phi^{\prime}}) = \mathtt{cross\_entropy}\left( Q_{\phi^{\prime}}(\mathbf{o}_t, l, \mathbf{s}_t, \mathbf{\epsilon}_t),  Q_{\phi}(\mathbf{o}_t, l, \mathbf{s}_t, \pi_{\theta}(\mathbf{o}_t, l, \mathbf{s}_t|\mathbf{\epsilon}_t))\right), \mathbf{\epsilon}_t\sim
    \begin{cases} 
        \mathcal{N}(\mathbf{0}, \mathbf{1}) & \text{\textit{w.p.} } 0.5, \\
        \pi_{\theta^{\prime}}(\mathbf{o}_t, l, \mathbf{s}_t) & \textrm{otherwise}.
        \end{cases}
\end{equation}
Different from the original implementation~\cite{wagenmaker2025steering}, to ensure a good coverage on the noise space when distilling $Q_{\phi^{\prime}}$, we sample the input noise from the original normal distribution with 0.5 probability (\textit{w.p.} 0.5), and from the noise predictor otherwise.

For sample-efficient offline to online adaptation, we maintain an off-policy buffer and an on-policy buffer, and sample batches from them evenly. Before training starts, we warm up the off-policy buffer with online rollouts of offline-trained checkpoints, similar to Warm-start RL~\cite{zhou2024efficient}. We intentionally choose not to mix teleoperated trajectories into the buffer to prevent the policy from training on mismatched dynamics. The on-policy buffer only stores trajectories generated from the two most recent checkpoints, and the stale data is pushed into the off-policy buffer.

\section{Robot \& System}
\label{sec:robot_system}

\begin{figure}
     \centering
     \includegraphics[width=0.9\linewidth]{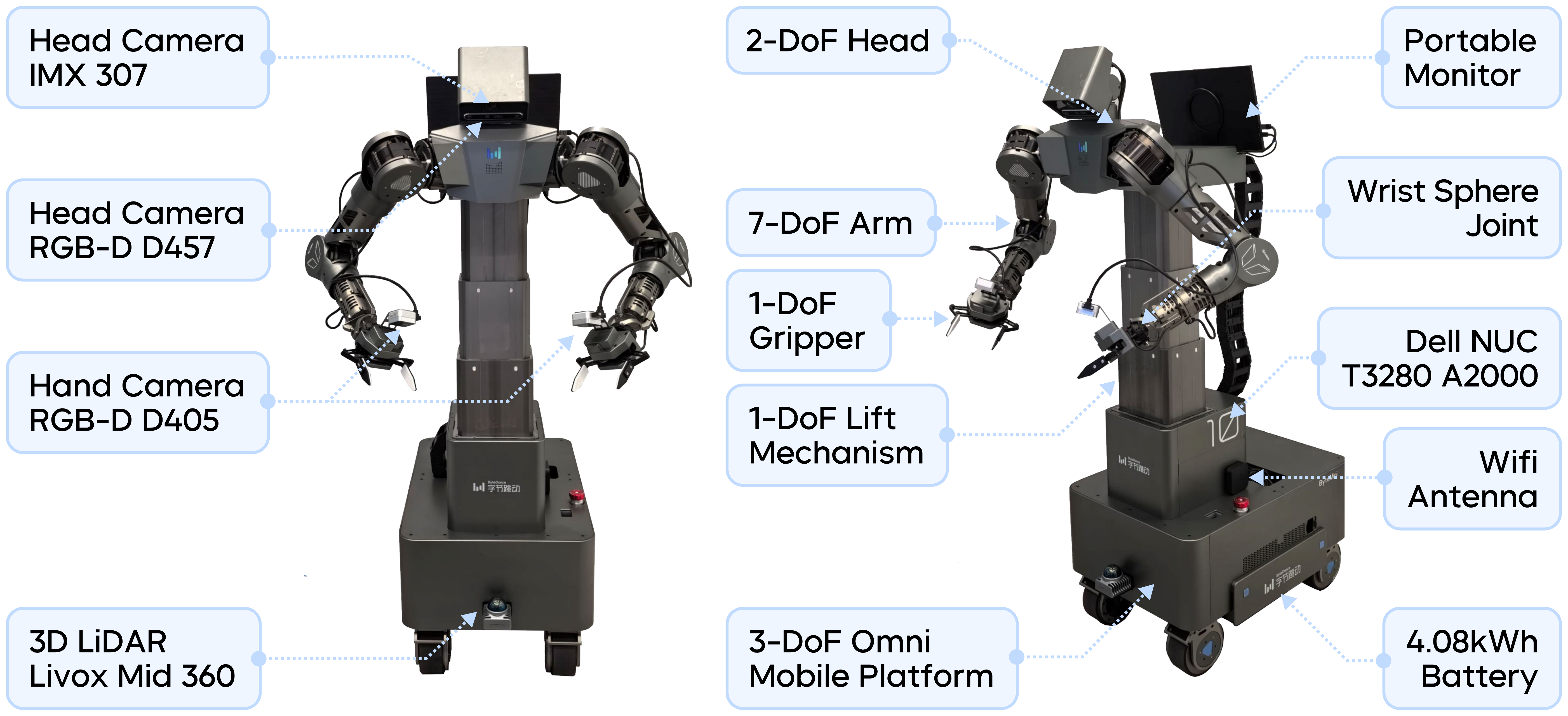}
     \caption{\textbf{The ByteMini-v2 Robot.} We show the robot specifications in terms of sensors, DoFs and electronic devices.}
     \label{fig:bytemini_robot}
 \end{figure}

The robot we utilize to verify GR-RL is \textbf{ByteMini-v2}, as illustrated in Fig.\,\ref{fig:bytemini_robot}. ByteMini-v2 is a wheeled mobile manipulation robot equipped with 7-DoF dual robotic arms, featuring a unique wrist spherical joint design~\cite{tian2025bytewrist}. The operational dexterity, stability, consistency, and usability of ByteMini-v1 robots have been fully validated in the work of GR-3~\cite{cheang2025gr}. Meanwhile, drawing on the limitations identified during ByteMini-v1 usage in GR-3, we have implemented the following three key design optimizations on ByteMini-v2 robots.

 \paragraph{Higher Load for Manipulation}
 The maximum load ability for the 7-DoF arm is limited by the maximum output torque of the elbow actuator. We update the elbow actuator from peak output torque 17\,Nm to 35\,Nm. As a result, the peak load of 7-DoF arms on ByteMini-v2 robots is increased from 1.4\,kg to 3.15\,kg, which will significantly expand the robot's operational capabilities.

\paragraph{Enhanced Mobility in Confined Spaces}
The projected area of the ByteMini-v1 chassis is 500\,mm x 720\,mm. And now for improving the mobility in confined spaces, the projected area of ByteMini-v2 robot is reduced to 450\,mm x 650\,mm respectively. Meanwhile, the design of the servo steering wheels on the mobile platform has also been optimized accordingly, enabling the synchronous adjustment of the steering wheels' motion in both yaw and pitch directions~\cite{wu2023tidybot} and thereby enhancing the robot's rapid direction-changing capability in confined spaces.

\paragraph{Higher Polished Robot Design and Enhanced Usability}
As shown in Fig.\,\ref{fig:bytemini_robot}, the ByteMini-v2 robot has been designed with enhanced refinement, featuring an ID and proper encapsulation of exposed electrical wiring harnesses. The position of the portable monitor has been adjusted from the robots' chassis to the shoulder, in order to achieve a better user experience. 
\section{Experiments}
\label{sec:experiments}

\paragraph{Task Description}
We present \name{} in a shoe lacing task, a challenging scenario featuring long-horizon, dexterous and precise manipulation. The observations are composed of three views of RGB images, proprioception states, and language instructions. 
In model inference, we incorporate a trajectory optimization module that imposes constraints on jerk and temporal continuity to refine the predicted action chunks. 
During RL training, we adopt a binary sparse reward setting, meaning that a positive reward of 1 is obtained only when the shoelace is threaded through the correct eyelet and put down on the table completely. 

\paragraph{Main Results}
\begin{figure}[!htb]
    \centering
    \includegraphics[width=0.8\linewidth]{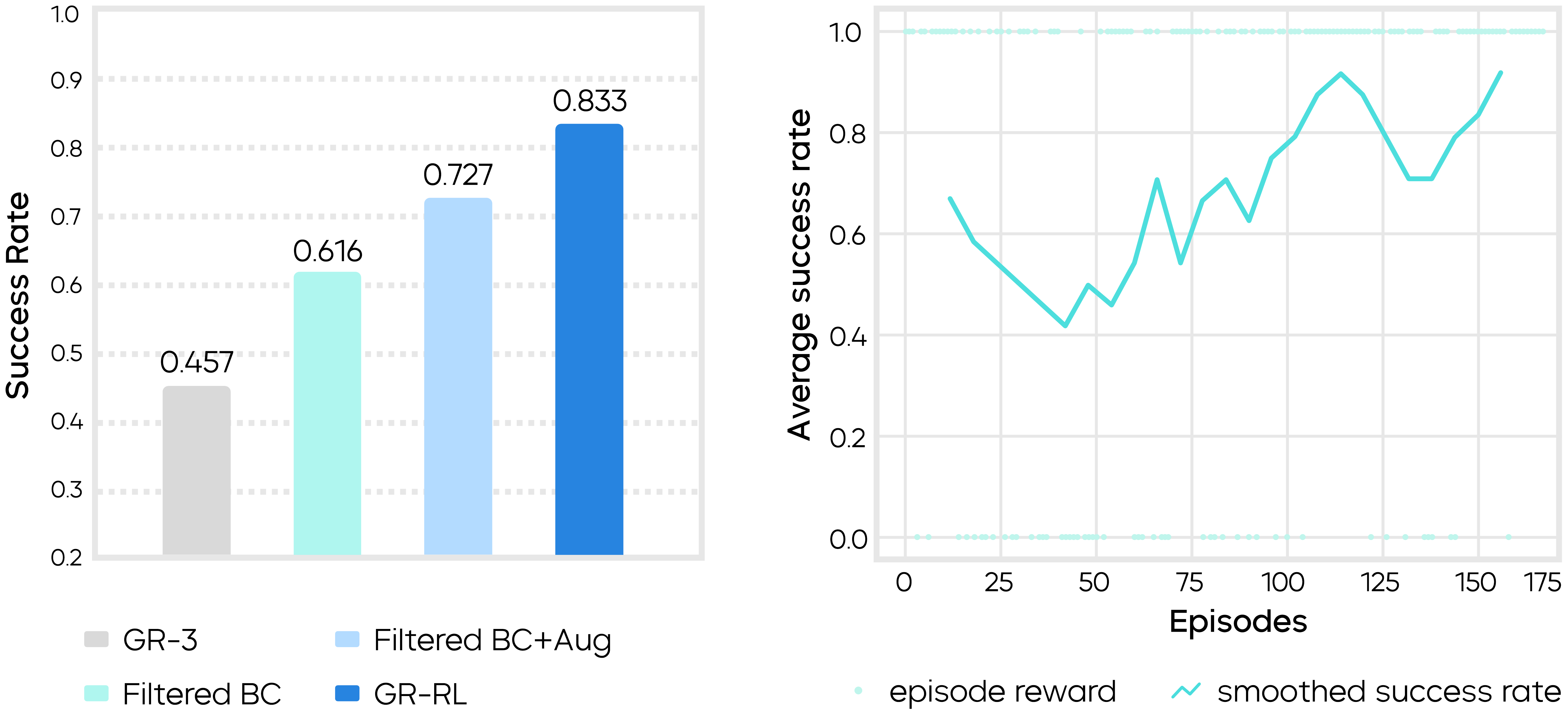}
    \caption{\textit{Left}: the success rate of our multi-stage training recipe. Data filtering, mirror augmentation, and online tuning all contribute to the final performance. \textit{Right}: the binary success signal per episode (dots) and the moving average of success rate (curve) during online finetuning. The performance increases rapidly after an offline-to-online adaptation phase.}
    \label{fig:exp:success_rate}
\end{figure}
We report the effect of our multi-stage training recipe in Fig.~\ref{fig:exp:success_rate}. Our base model GR-3, trained with behavior cloning over all the human teleoperated data, achieves a success rate of 45.7\%. After filtering the data with our learned task progress model, the success rate is boosted to 61.6\%. It highlights the importance of the data cleaning mechanism when learning precise and long-horizon manipulation. With symmetry data augmentation, the success rate of a filtered behavior cloning policy is further increased to 72.7\%.

\begin{figure}[t]
    \centering
    \includegraphics[width=0.8\linewidth]{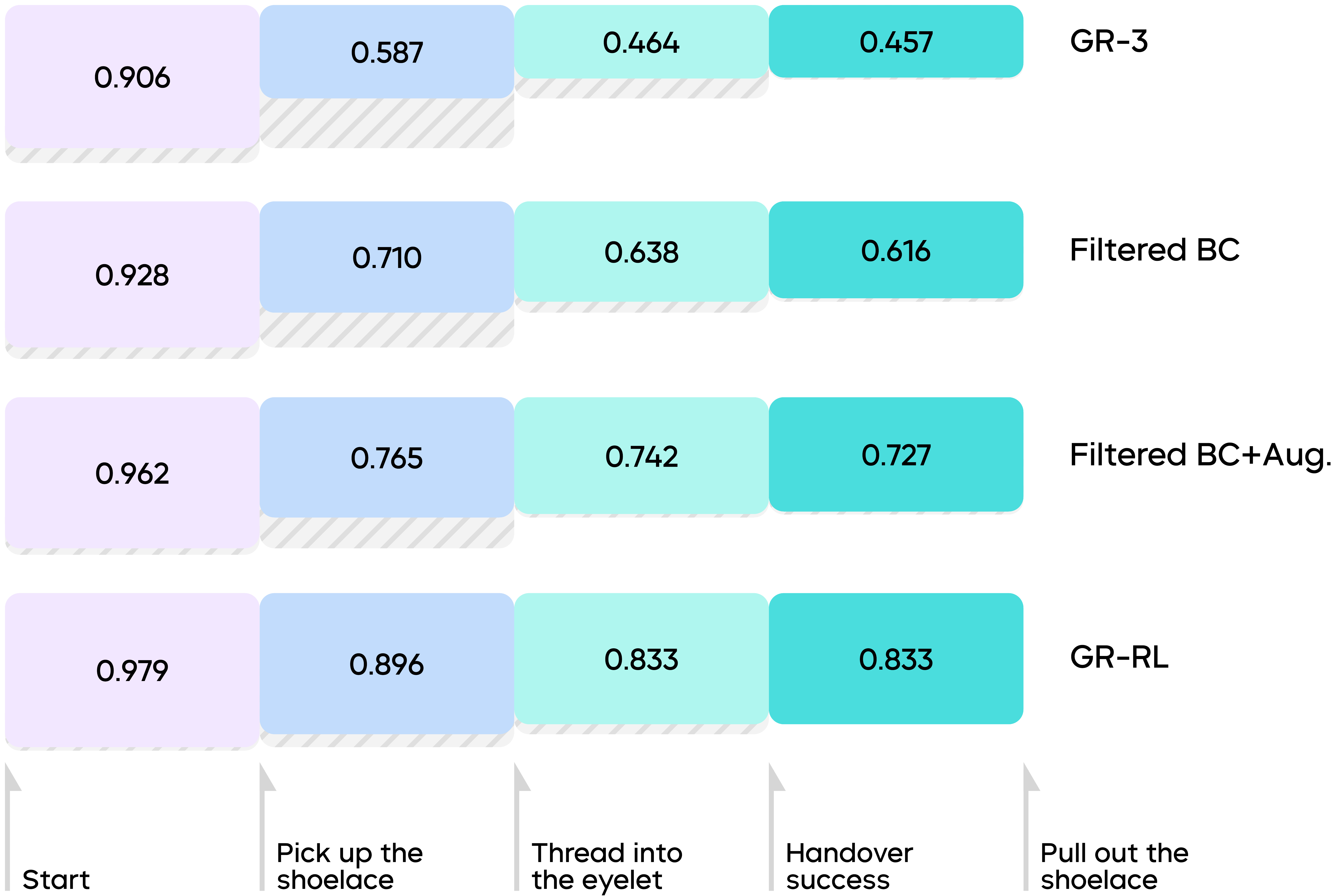}
    \caption{Detailed success rates of different models for completing intermediate stages. The height of each hatched area denotes the decrease in success rate from the previous stage to the current stage.}
    \label{fig:exp:stage_success}
\end{figure}

The ``Filtered BC + Aug.'' model serves as the starting point of online steering RL. To better align the offline-trained critic to the distribution of model rollout, we collect 673 trajectories generated by the offline model, then freeze the VLM backbone and continue to train the critic heads $Q_{\phi}$, $Q_{\phi^{\prime}}$ using these data. These rollout trajectories are also populated into the off-policy replay buffer in online RL to stabilize training. We then tune both the noise predictor $\pi_{\theta^{\prime}}$ and the two critic heads for 50 optimization steps once we gather 12 new episodes online. The moving average of the success rate over a window size of 24 is shown as the curve on the right side of Fig.~\ref{fig:exp:success_rate}, and the binary success signals for all online episodes are represented as dots. In the first few iterations, the success rate decreases due to the distribution shift from offline to online RL. Later, the success rate rapidly recovers and grows beyond the offline performance to over 90\%. The checkpoint at 500 online training steps is used for evaluation, and achieves a success rate of 83.3\%.

To better understand the failure modes for different models in this long-horizon task, we evaluate whether the models succeed at several critical stages, including picking up the correct shoelace, threading the shoelace into the correct eyelet, handing it over to another gripper, and pulling the shoelace tight. The detailed success rate is illustrated in Fig.~\ref{fig:exp:stage_success}. The colored area denotes the success rate for completing each stage. The hatched area represents the decrease in the success rate compared to the previous stage. Data filtering and online RL can largely reduce the failure during threading. Data augmentation improves the model performance in all stages, although with a smaller magnitude.

\paragraph{Ablation on the Progress Evaluator}
We validate the effectiveness of the RL-based progress evaluator by comparing it with a regression variant. The regression baseline is trained by directly regressing the temporal progress $\frac{t}{T}$ of each state $\mathbf{s}_t$ in successful trajectories.
As shown in Fig.~\ref{fig:method:progress}, the regression-based predictor tends to overly smooth the progress prediction. It makes reasonable predictions in normal cases but is less sensitive to subtle failure (oftentimes only millimeters from success), such as failing to pull out the shoelace or imprecise insertion. Also, the regression-based predictor is not good at capturing transitions with long-term effects. There is a significant value jump in the predictions of GR-RL when the robot intentionally puts down the shoelace to adjust the grasping pose, but the regression-based prediction is almost flat during the adjustment.

\begin{figure}[!htb]
    \centering
    \includegraphics[width=0.9\linewidth]{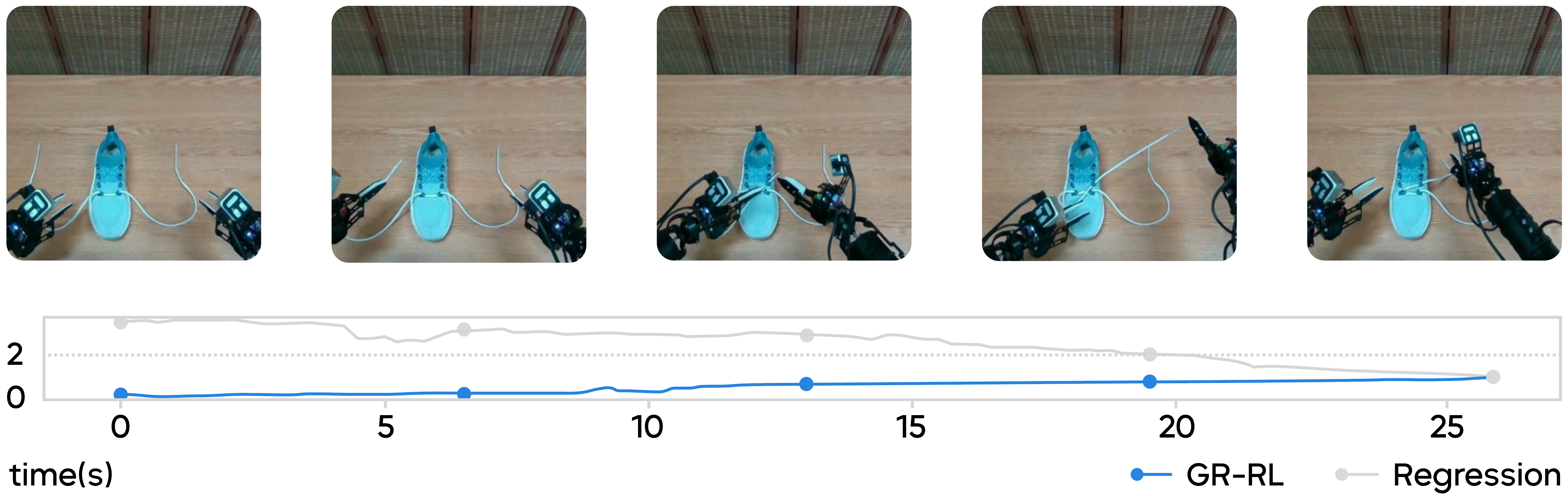}
    \caption{Comparison of progress prediction by distributional vs. non-distributional critics. Non-distributional critics are unbounded in the output range, and fail to reflect the positive progress in a successful trajectory.}
    \label{fig:exp:progress_distributional}
\end{figure}

We also train a non-distributional critic with the same TD3+BC algorithm to compare with our distributional critic. The models are evaluated over a high-quality successful trajectory, as illustrated in Fig.~\ref{fig:exp:progress_distributional}. Due to the long horizon and binary sparse reward in our setting, the non-distributional critic suffers from severe over-estimation, especially in earlier parts of trajectories where the reward supervision signal is weak. The value prediction of our distributional critic falls in a predefined range, thus converging to a reasonable scale more robustly and demonstrating better alignment with the true temporal order. 

\begin{figure}
    \centering
    \begin{subfigure}[b]{\textwidth}
        \centering
        \includegraphics[width=\linewidth]{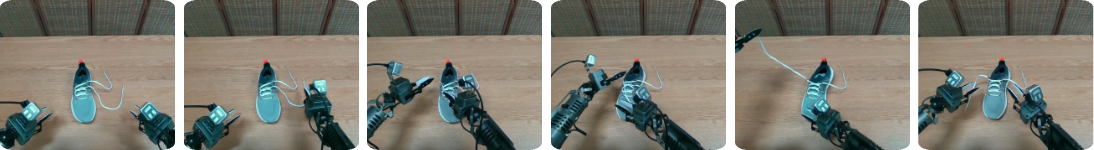}
        \caption{Thread shoelace for a shoe with a different color.}\label{fig:exp:case_color}
    \end{subfigure}
    \begin{subfigure}[b]{\textwidth}
        \centering
        \includegraphics[width=\linewidth]{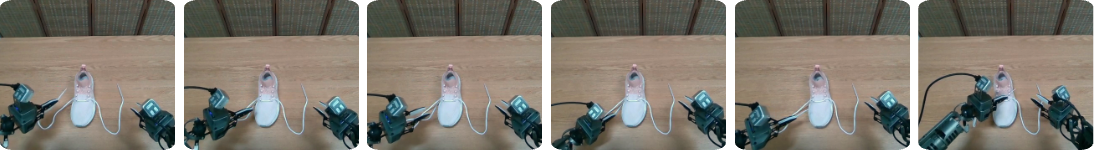}
        \caption{Regrasp the shoelace when it drops.}\label{fig:exp:case_regrasp}
    \end{subfigure}
    \begin{subfigure}[b]{\textwidth}
        \centering
        \includegraphics[width=\linewidth]{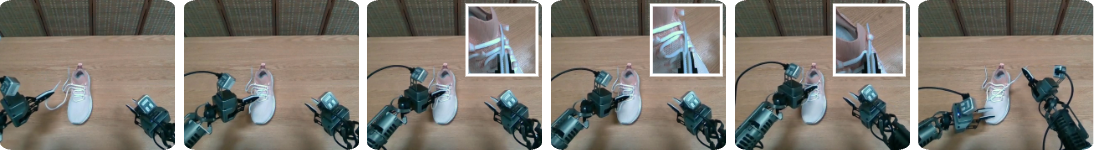}
        \caption{Re-attempt when the shoelace is not threaded precisely through the eyelet.}\label{fig:exp:case_thread_retry}
    \end{subfigure}
    \begin{subfigure}[b]{\textwidth}
        \centering
        \includegraphics[width=\linewidth]{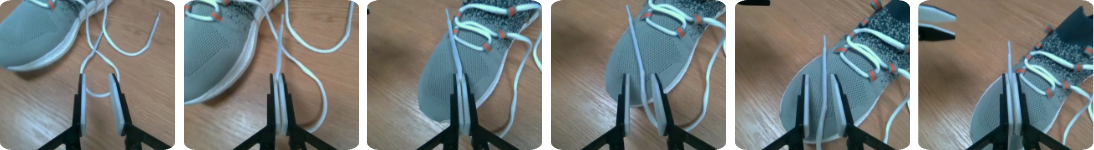}
        \caption{Adjust the grasp pose intentionally on the surface of the shoe.}\label{fig:exp:case_adjust_grasp}
    \end{subfigure}
    
    \begin{subfigure}[b]{\textwidth}
        \centering
        \includegraphics[width=\linewidth]{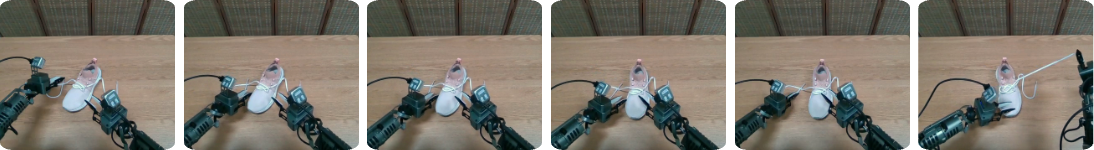}
        \caption{Reorient the shoe before threading.}\label{fig:exp:case_reorient_shoe}
    \end{subfigure}

    \begin{subfigure}[b]{\textwidth}
        \centering
        \includegraphics[width=\linewidth]{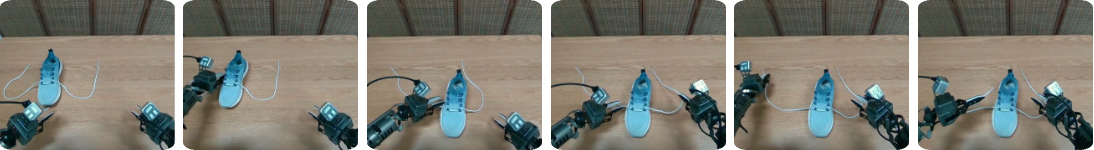}
        \caption{Adjust the positions of the shoe and the shoelace before threading.}\label{fig:exp:case_shoe_pos}
    \end{subfigure}

    \begin{subfigure}[b]{\textwidth}
        \centering
        \includegraphics[width=\linewidth]{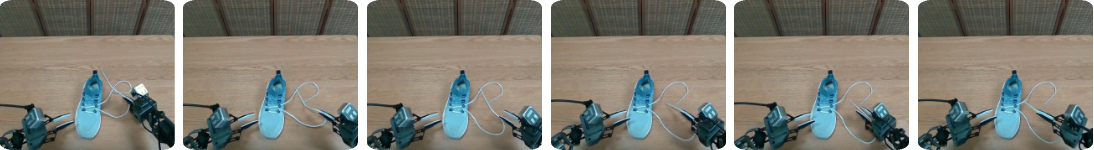}
        \caption{Pull out the shoelace underneath another one from the crossed laces.}\label{fig:exp:case_cross_shoelace}
    \end{subfigure}
    
    \caption{Robust behavior of \name{} in various cases.}
    \label{fig:exp:case_study}
\end{figure}

\paragraph{Visualization of the Learned Behaviors}
\name{} demonstrates robust behaviors in various cases. It can handle shoes with different colors and sizes, as shown in Fig.~\ref{fig:exp:case_study}. The model automatically retries when the shoelace accidentally drops down (Fig.~\ref{fig:exp:case_regrasp}) or when the shoelace misses the eyelet (Fig.~\ref{fig:exp:case_thread_retry}). The model can actively manipulate the scene to make the task easier to solve. In Fig.~\ref{fig:exp:case_adjust_grasp}, the initial grasping point is far from the tip of the shoelace. The model decides to drop the shoelace on top of the deformable shoe and regrasp closer to the tip before threading. In Fig.~\ref{fig:exp:case_reorient_shoe}, the model first reorients the shoe from the left side to straighten it, then starts threading. Similarly, for a shoe initially positioned on the far side of the table (Fig.~\ref{fig:exp:case_shoe_pos}), the robot can pull it near, adjust the position of the shoelace, and then complete the task. In cases where the two ends of the shoelace are crossed, and the end that should be grasped is underneath (Fig.~\ref{fig:exp:case_cross_shoelace}), the model can identify the correct one and pull it out.

\section{Related Work}

\paragraph{Generalist Robotic Foundation Policy}
Building generalist robotic foundation manipulation policies is a long-standing challenge for robotics research and applications~\cite{brohan2022rt, brohan2023rt, kimopenvla, black2410pi0, intelligence2025pi_, bjorck2025gr00t, team2024octo, li2023vision, li2024towards, reuss2024multimodal, bharadhwaj2024roboagent, shukor2025smolvla, barreiros2025careful}.
Recent advances in building vision-language-action (VLA) models advocate adapting the vision-language models (VLMs) pretrained with web-scale data to robotic actions by adding the action modality~\cite{o2024open, team2024octo, kimopenvla, black2410pi0, pertsch2025fast, intelligence2025pi_, doshi2024scaling, wang2024scaling, liu2024rdt, qu2025spatialvla, li2024cogact}.
The core idea is to leverage large-scale real-world robot trajectories collected by human teleoperators and generalize to potentially novel scenes and tasks.
\name{} sits upon the prior success of GR-3~\cite{cheang2025gr}, a generalist policy co-trained with web-scale data and human demonstrations. \name{} further improves GR-3 by filtering high-quality data, augmenting actions, and online real-world RL, enabling it to perform long-horizon dexterous and precise manipulation.

\paragraph{Real-World Reinforcement Learning}
A central limitation of pure imitation learning is its susceptibility to compounding errors and its inability to exceed the performance of demonstrations. To address these issues, a significant body of work explores online data collection and real-world reinforcement learning (RL) to improve manipulation robustness beyond supervised training~\cite{li2025reinforcement,levine2016end,lv2025flow,luo2024serl,ankile2025residual,sharma2023self,kalashnikov2018scalable,seo2024continuous,luo2024precise}. 
In the context of VLAs, some recent works aim to perform policy improvement via on-policy RL~\cite{schulman2017proximal,lu2025vla,liu2025can,chen2025pi_,xia2025robotic} in simulation. However, transferring their success to real-world scenarios remains difficult because real-world interactions are sample inefficient and noisy. 
Another line of work learns world models and performs on-policy RL interacting with the learned world model~\cite{hafner2019dream,hafner2020mastering,hafner2023mastering,wu2023daydreamer,hansen2022temporal,hansen2023td,zhu2025wmpo,ma2021contrastive}. World models alleviate the issue for real-robot interaction, but introduce further issues given inaccurate visual predictions. \name{} follows~\cite{luo2024serl,seo2024continuous,luo2024precise,xiao2025self} and focuses on real-world off-policy RL. This allows us to efficiently utilize past trajectories and improve the sample efficiency. 
Treating the noisy real-world reward as a distribution, we show that distributional critics significantly improve the robustness compared with standard regression-based critic models.\
Concurrent to our work, $\pi^*_{0.6}$ presents a real-world RL pipeline for high-precision manipulation~\cite{physical2025pi06}. Similar to $\pi^*_{0.6}$, we both adopt distributional critics that learn the progress of the task. However, instead of performing advantage-conditioned denoising, we directly perform filtered behavior cloning and also observe a strong performance boost.
Given the stronger base offline policy, it helps reduce the search space during online exploration.
We hope \name{} reveals certain insights to the community on building capable specialist agents from generalist policies and helps to push the boundaries of deployable robotics research.

\section{Limitations \& Conclusions}
\label{sec:conclusions}

\paragraph{Limitations}
Despite the strong performance of \name{} in long-horizon high-precision dexterous tasks, it still has clear limitations. 
One of the major issues of our current pipeline is the behavior-drifting problem. Given a sparse and noisy reward, our policy behavior could be unstable during online RL. This is possibly due to the limited capacity of the lightweight noise predictor, or the challenging credit assignment issue in the large latent action space.
Furthermore, distilling the improved policy into the base VLA could be a potential direction for obtaining both capable and general manipulation policies.
We leave these issues for future study.

\paragraph{Conclusion}
We introduce \name{}, a robotic learning framework for building specialist VLA policies for long-horizon dexterous and precise manipulation. The key insight of \name{} is that the mismatch between data collection and policy inference needs online alignment.
\name{} learns an RL-based evaluator by treating the critic value learned from sparse reward as task progress prediction, and uses it to filter high-quality transitions to train a robust base policy. During this process, we also introduce a simple yet effective morphological symmetry augmentation method to improve the overall performance. Last, we perform online RL that aligns the rollout behavior with our training signals.
To the best of our knowledge, \name{} is the first learning-based policy capable of lacing up shoes. We hope \name{} can be a small step towards capable real-world specialist robot policies.

\clearpage

\bibliographystyle{plainnat}
\bibliography{main}
\clearpage
\section*{Contributions and Acknowledgments}
\phantomsection
\addcontentsline{toc}{section}{Contributions and Acknowledgments}
\label{sec:contributions}
Authors are listed in alphabetical order.

\paragraph{Core Contributors} Yunfei Li, Xiao Ma, Jiafeng Xu

\paragraph{Contributors} Yu Cui, Zhongren Cui, Zhigang Han, Liqun Huang, Tao Kong, Yuxiao Liu, Hao Niu, Wanli Peng, Jingchao Qiao, Zeyu Ren, Haixin Shi, Zhi Su, Jiawen Tian, Yuyang Xiao, Shenyu Zhang, Liwei Zheng

\paragraph{Supervisors} Hang Li, Yonghui Wu

\paragraph{Acknowledgments} We would like to thank Yichu Yang for the insightful discussion on morphological mirror augmentation. We thank Chilam Cheang, Jinming Guo, Zetian Li, Xin Zhao, Mingyang Wang, Zhiguo Hao, Tianxiang Gong, Yang Zhao, Shuzhai Guo, Ziye Liu and all the data annotators for their help on data curation. We thank Degong Yang and Yang Liu for their help on hardware system development and maintenance.

\end{document}